\documentclass[a4paper,twoside]{article}

\usepackage{epsfig}
\usepackage{calc}
\usepackage{amssymb}
\usepackage{amstext}
\usepackage{amsmath}
\usepackage{amsthm}
\usepackage{multicol}
\usepackage{pslatex}
\usepackage{apalike}
\usepackage{booktabs}
\usepackage{algorithm}
\usepackage{algorithmic}
\usepackage{marvosym}
\usepackage{scrextend}
\usepackage{gensymb}
\usepackage{subfig}
\usepackage{array,multirow,graphicx}
\usepackage{amsfonts}
\usepackage{bm}
\usepackage{times}
\usepackage{soul}
\usepackage{url}
\usepackage[hidelinks]{hyperref}
\usepackage[utf8]{inputenc}
\usepackage{SCITEPRESS}     

\hypersetup{
    colorlinks=true,
    linkcolor=black,
    filecolor=magenta,      
    urlcolor=blue,
    citecolor=black
}

\begin{document}

\title{Explorations and Lessons Learned in Building an Autonomous Formula SAE\thanks{A student formula design competition organized by SAE International (previously known as the Society of Automotive Engineers).} Car from Simulations}

\author{\authorname{Dean Zadok\thanks{The authors contributed equally to this work.}\sup{1}, Tom Hirshberg\footnotemark[\value{footnote}]\sup{1}, Amir Biran\sup{1}, Kira Radinsky\sup{1} and Ashish Kapoor\sup{2}}
\affiliation{\sup{1}Computer Science Department, Technion, Haifa, Israel}
\affiliation{\sup{2}Microsoft Research, Redmond, WA}
\email{\{deanzadok, tomhirshberg, amirbir\}@campus.technion.ac.il, kirar@cs.technion.ac.il, akapoor@microsoft.com}
}

\keywords{Autonomous Driving, Imitation Learning, Deep Neural Networks}

\abstract{This paper describes the exploration and learnings during the process of developing a self-driving algorithm in simulation, followed by deployment on a real car. We specifically concentrate on the Formula Student Driverless competition. In such competitions, a formula race car, designed and built by students, is challenged to drive through previously unseen tracks that are marked by traffic cones. We explore and highlight the challenges associated with training a deep neural network that uses a single camera as input for inferring car steering angles in real-time. The paper explores in-depth creation of simulation, usage of simulations to train and validate the software stack and then finally the engineering challenges associated with the deployment of the system in real-world.}

\onecolumn \maketitle \normalsize \setcounter{footnote}{0} \vfill

\section{\uppercase{Introduction}}
\label{sec:introduction}

\noindent Machine Learning (ML) paradigms such as supervised learning, imitation learning, learning-by-demonstration and transfer learning are significantly transforming the field of robotics. However, training a real-world robot is quite challenging due to high sample complexity of these methods - successful training of deep Neural Networks require millions of examples. The primary means to solve this problem has been to train models in simulation and then deploy them in the real world. There are various challenges with such a workflow \cite{QuadS2R,RoboticS2R} and this paper explores these issues by creating a real-world autonomous car that was trained completely end-to-end in a simulation.

In particular, the objective of this work was to create an autonomous formula student car that could compete in an international self-driving formula student competition in Germany. The competition takes place on a professional Formula racetrack. Two lines of traffic cones (blue and yellow cones) define the track. Note that the track is unknown prior to the race, so the goal was to create an autonomous system that could maneuver around accurately on arbitrary and unfamiliar tracks. Further, the race scene could potentially include ad signs, tires, grandstands, fences, etc.

\begin{figure}[t]
\begin{center}
\begin{tabular}{>{\centering\arraybackslash} m{1.5cm} >{\centering\arraybackslash} m{3.2cm} >{\centering\arraybackslash} m{1.5cm}}
\subfloat{\includegraphics[width = 0.095\textwidth]{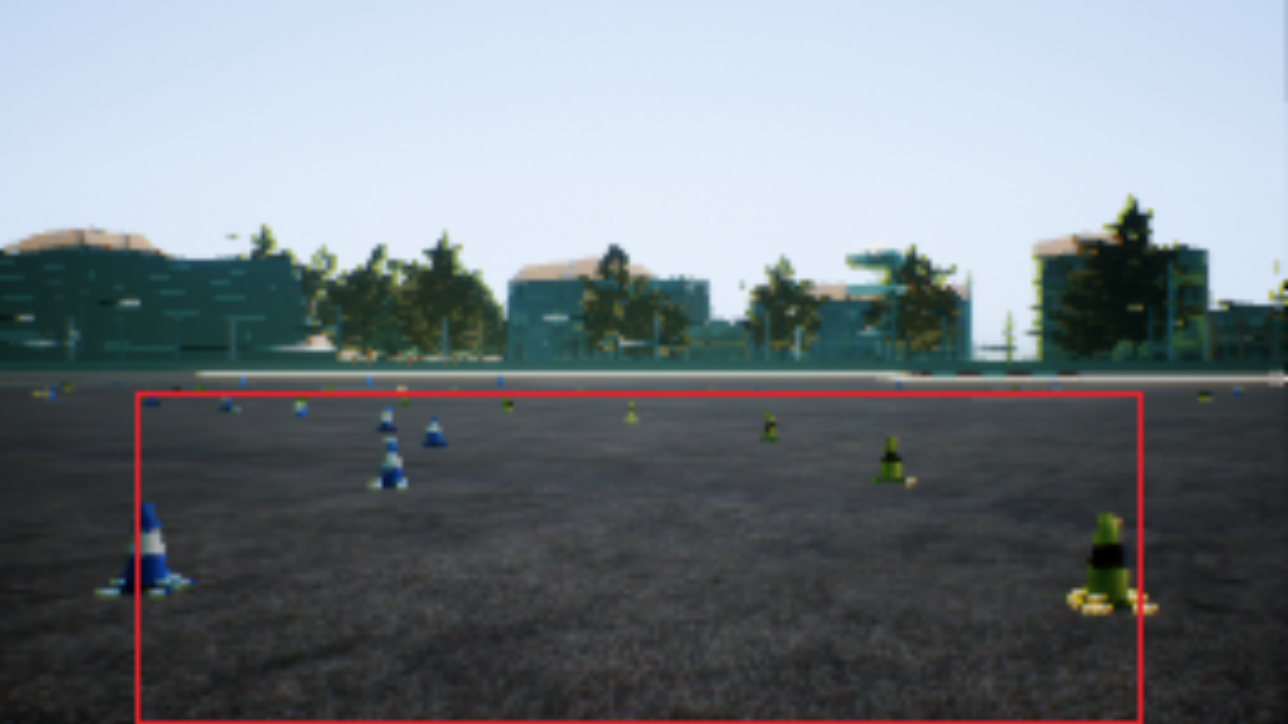}} & \subfloat{\includegraphics[width = 0.203\textwidth]{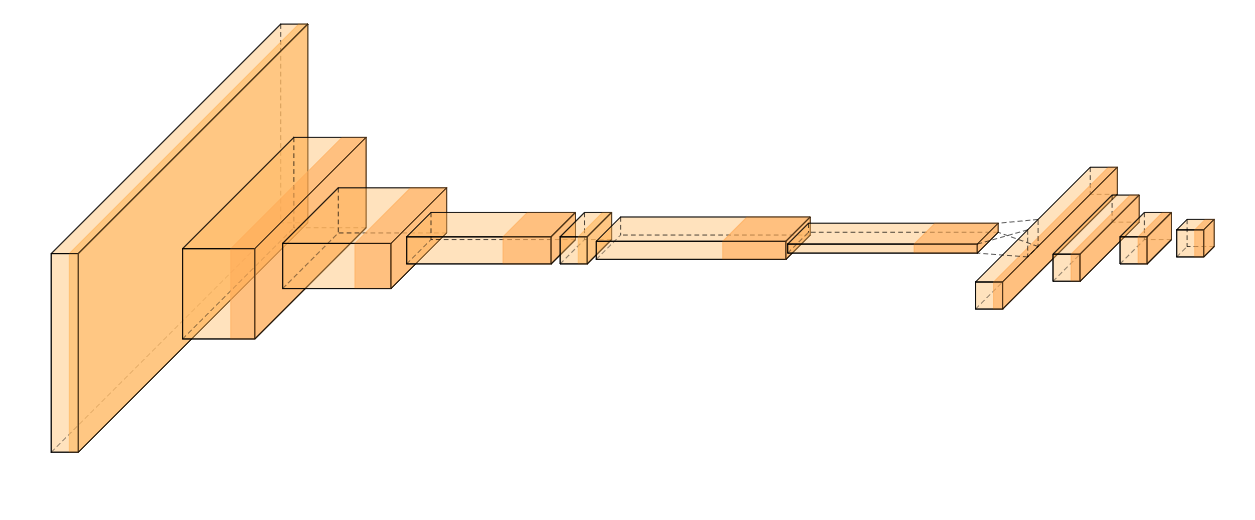}} & \subfloat{\includegraphics[width = 0.095\textwidth]{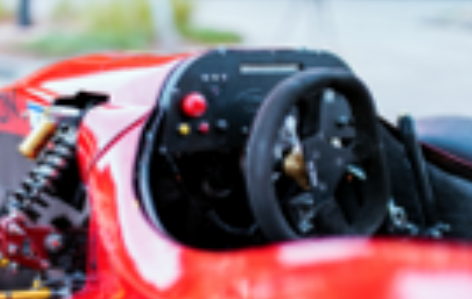}} \\[-1.8ex]
{\tiny Input} & {\tiny DNN} & {\tiny Controller}
\end{tabular}
\end{center}
\caption{End-to-end deep learning approach. The controller acts according to a predicted steering angle from the trained DNN. The predicted steering angle is inferred from a region of interest of a given image. This methodology is applied both in simulation, and in real-world environments.}
\label{figure:diagram}
\end{figure}

\begin{figure*}[!ht]
\centering
\subfloat{\includegraphics[width = 0.245\textwidth]{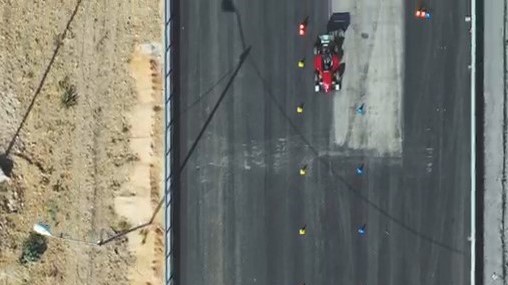}} 
\hspace{0.02cm}
\subfloat{\includegraphics[width = 0.245\textwidth]{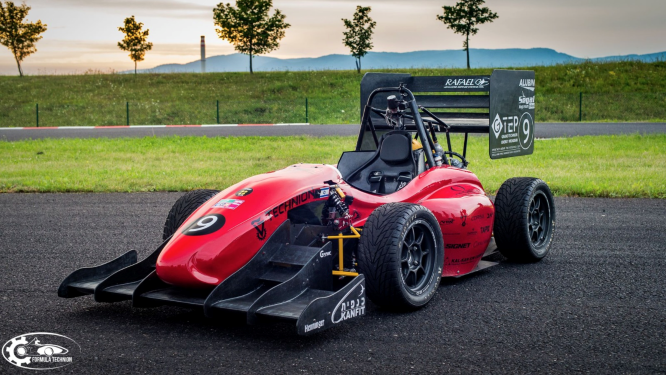}}
\hspace{0.02cm}
\subfloat{\includegraphics[width = 0.245\textwidth]{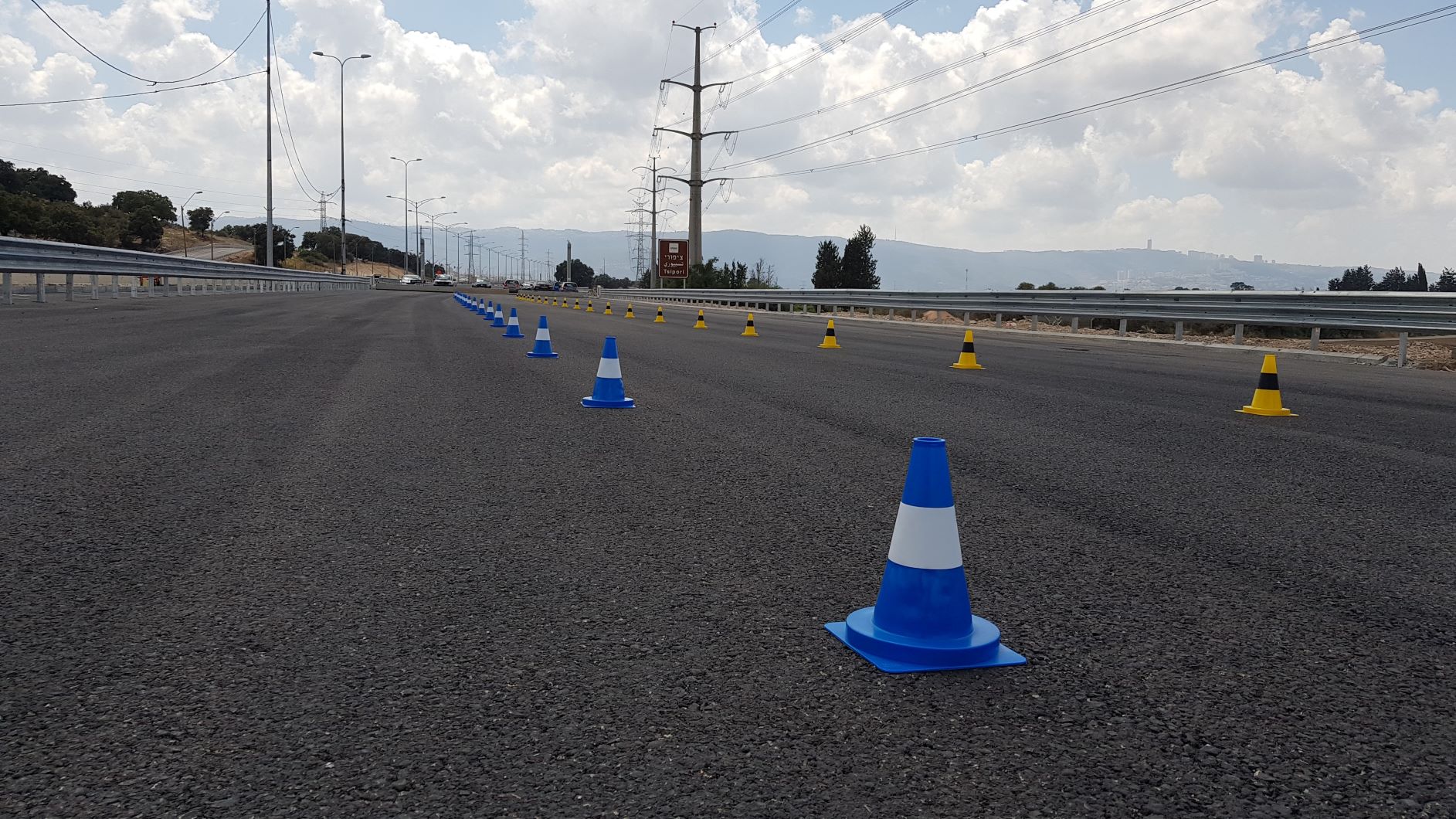}}
\hspace{0.02cm}
\subfloat{\includegraphics[width = 0.245\textwidth]{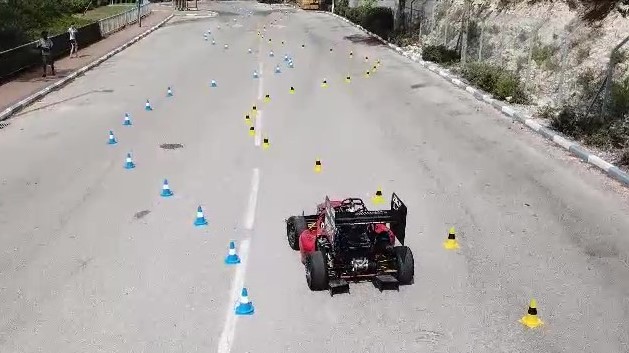}} \\ [-1.85ex]
\subfloat{\includegraphics[width = 0.245\textwidth]{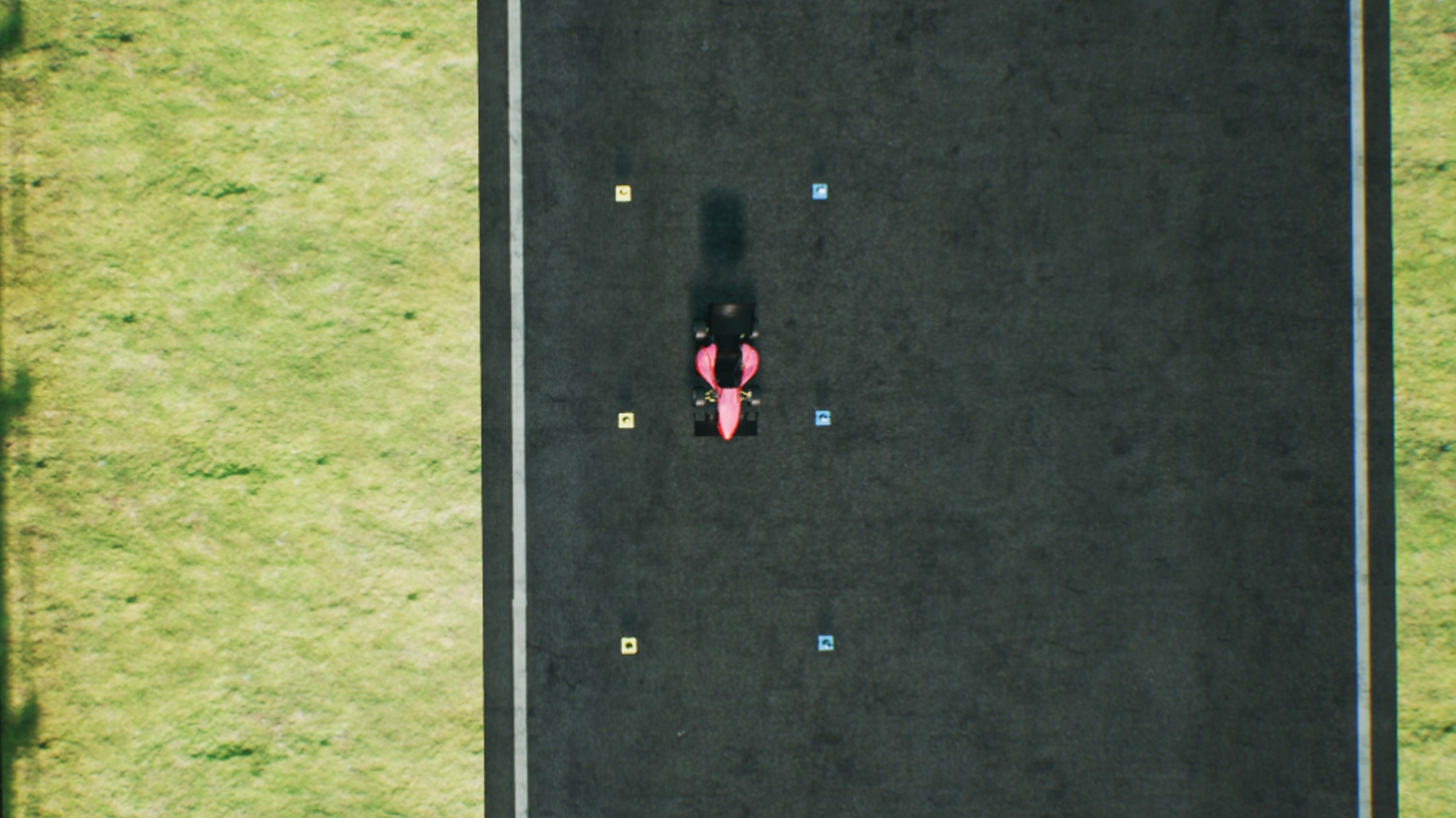}} 
\hspace{0.02cm}
\subfloat{\includegraphics[width = 0.245\textwidth]{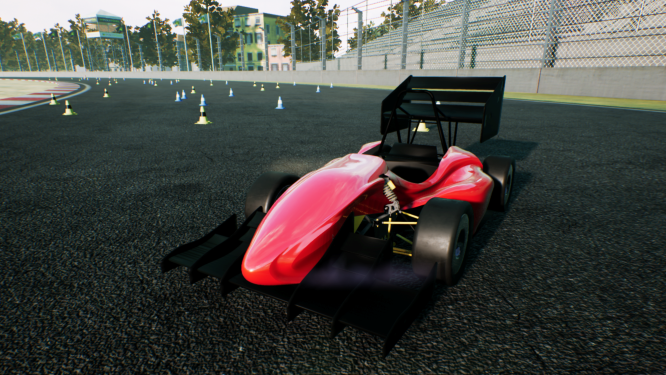}} 
\hspace{0.02cm}
\subfloat{\includegraphics[width = 0.245\textwidth]{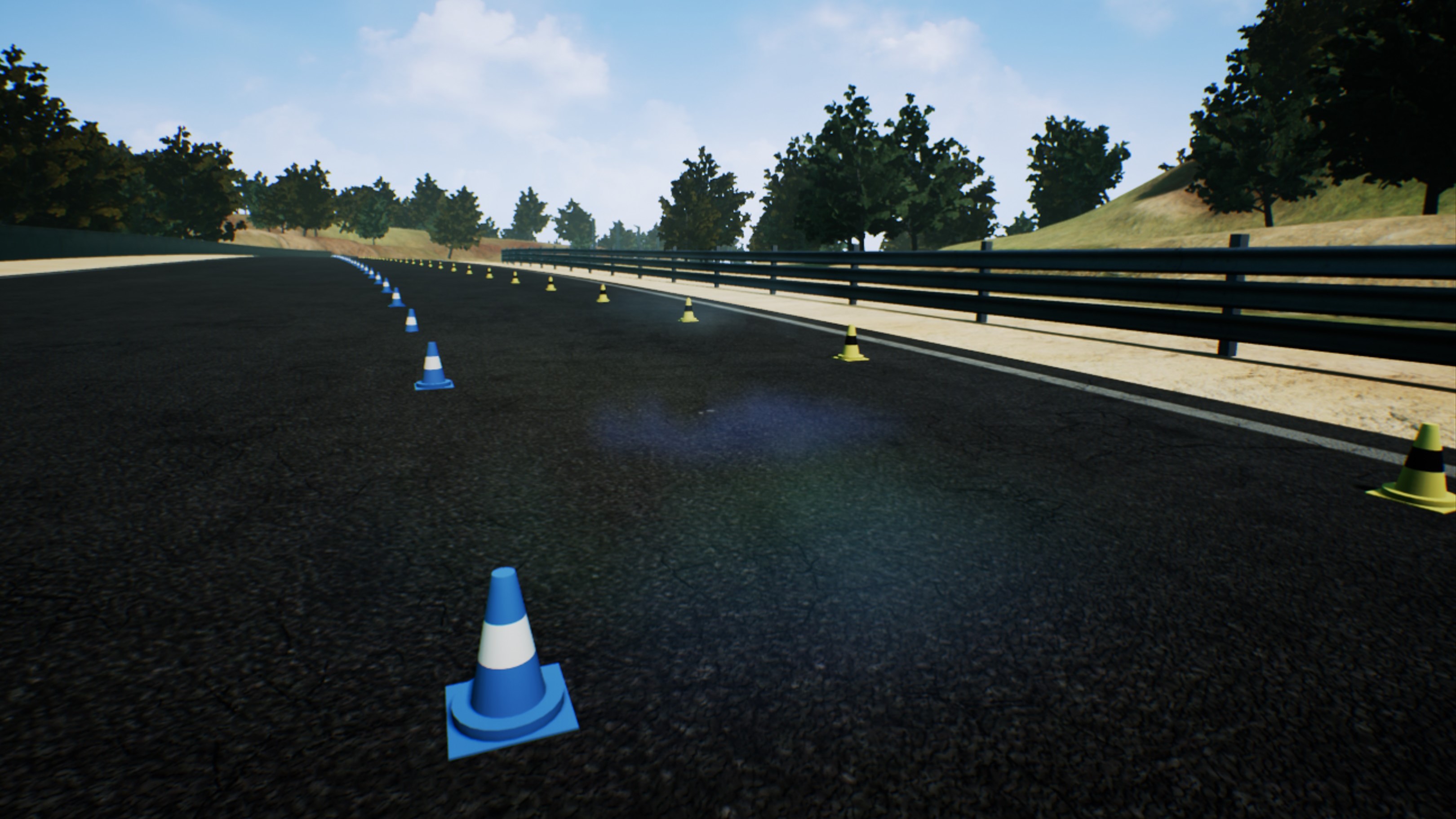}} 
\hspace{0.02cm}
\subfloat{\includegraphics[width = 0.245\textwidth]{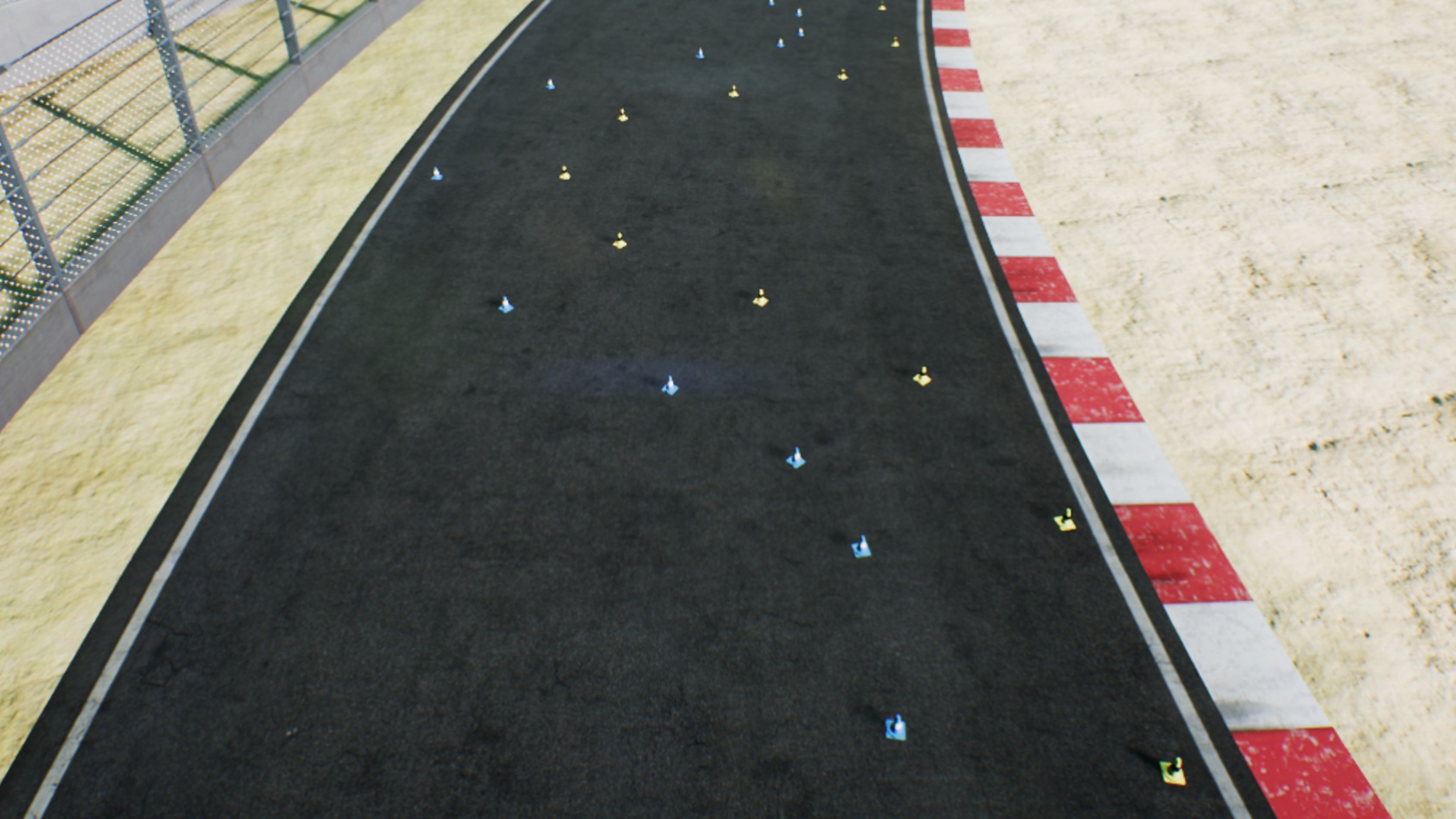}} 
\caption{Images taken from our simulated environment (bottom) and from real tracks we have driven in (top). The simulation was built to have objects resemble their real-world counterparts precisely, including the exact traffic cones and our Formula SAE car. Both the real world and the simulated environments consisted of straight and swerved tracks as shown. Our model was trained in simulation only and was executed in a real-world environment.}
\label{figure:collage}
\end{figure*}

There are several challenges associated with designing and implementing such a system. First, creating a test environment in order to engineer such a system is not only cumbersome and expensive, but also quite complex due to safety considerations. Secondly, building an end-to-end system requires several complex engineered systems, both in hardware and software, to work together - often such integrations are challenging, and such dependencies can greatly impede rapid development of individual systems (e.g., progress on autonomous software stack crucially depends upon bug-free hardware). Finally, much of the recent autonomous system modules depend upon the ability to collect a large amount of training data. Such data collection is not scalable in real-world due to both the complexity as well as resource requirements to carry out such training missions.

We alleviated these problems by training and validating our autonomous stack in simulation. Such simulations allow us to mitigate risks associated with safety, enable us to bypass the need to access a closed off-road track and minimize the dependency on the development of the hardware stack. Most importantly, such simulations allow us to gather data at scale, in a wide variety of conditions, which is not feasible in real-world. Our simulations consist of a dynamic model of the custom car, a variety of racetrack scenes and synthesis of different weather conditions. We explore and demonstrate that such simulations can indeed be very useful in building autonomous systems that are deployed in real-world. We especially highlight the importance of creating high-fidelity scenes that mimic the real-environment. We also aim to build our models completely in simulation, without requiring the system ever to see the real tracks.

At the heart of the system is a Deep Neural Network (DNN) which is trained via imitation learning \cite{codevilla2018end}. The main task of the DNN is to predict the desired set of actions the car should take in real time, given the state the car was in relative to the environment. The DNN was trained to imitate the actions of test subjects who first drove the car around in the simulation during the training data collection phase. We also investigate challenges associated with deploying such a trained model on a real car. Specifically, transferring the trained model from simulation to the real-world consisted of creating a computationally efficient deployment on an edge device that could result in a high frame-rate, for real-time predictive performance. This part of our work required various optimizations from software engineering including the ability to operate using low-power systems. In summary, various contributions of this paper include:

\begin{itemize}
\item An end-to-end design and deployment of an autonomous stack that can drive a custom Formula SAE car.
\item Unique augmentations that alleviate the recording procedure and improve the trained model substantially. 
\item A detailed overview of how such predictive systems trained in simulations can be deployed on a system that operates in a real-world environment.
\end{itemize}

The implementation is available at the repository

\url{https://github.com/FSTDriverless/AirSim}

\section{\uppercase{Related Work}}
\label{sec:RelatedWork}

\noindent The industry of autonomous driving has seen a lot of changes over the past few years. The improvements in computer vision applications using deep learning tools had encouraged teams to turn to modern methods rather than using classic algorithms, or to have incorporated these new technologies in hybrid-type approaches.
The DARPA competition, initiated in 2004, served as a ground for teams to advance autonomous technologies in a significant manner. Such team describes in \cite{darpa} the vehicle that had won the 2006 "Urban Challenge" - a part of the DARPA competition that year. The mechanism they used was composed of a perception system, and of three algorithmic components - mission, behavioral, and motion planning. The model they used, as was common in autonomous driving algorithms at the time, attempted to deal with the complexity of the task by modularizing the decision process and addressing each module seperately - as opposed to an end-to-end approach as described by our work.

As shown by the above example of the DARPA winner, autonomous driving common practices include separately addressing two stages of the process, one being perception and optional world model maintenance, and the other being the decision making process. The perception stage includes localization, which is often addressed using either Lidar \cite{RobustVL}, a camera \cite{BRU16}, or a combination of both\cite{XU17}. Perception may also include offline obstacle mapping such as in SLAM\cite{slam}, which will be discussed further on. Given the above sensor data, the self-driving system implements an object detection mechanism, that is usually based on deep-learning methods \cite{Peili,BRavi}. The decision making process involves analysis and inference over the internal world model, or directly over raw sensor data. Many classical graph-based trajectory planning methods are still being improved and incorporated into hybrid models \cite{BAS15}. Opposed to graph-based policy algorithms we find end-to-end approaches which use neural network image processing (e.g., bounding box object detection as in YOLO \cite{YOLO} and semantic segmentation \cite{DEEPLAB}) to infer policies from data which can be either raw or processed sensor input \cite{E2E1,E2E2}.

In 2017, the first Formula Student Driverless (FSD) competition took place, with only four teams passing regulations for full participation. Since it was the first time the competition was held, most of the teams took conservative approaches by implementing algorithms which are based on classic methods, and were attempting to get through the race track carefully and steadily. The system that won the first FSD competition is called SLAM \cite{slam}. In this method, the vehicle drives very slowly during the first lap. This is done in order to generate an accurate internal model of the track, so that in the next laps the car can drive through the track faster and navigate by maneuvering between obstacles without using visual sensors or high-computation demanding inference algorithms.

Sim-to-real approaches have also been extensively researched in recent years \cite{S2R1,S2R2}. Such models attempt at overcoming differences between the real world and simulation and benefiting from the endless amount and variation of data attainable on simulation, when used properly. Sim-to-real was used to train quad-copters \cite{QuadS2R}, mechanical robotics control \cite{RoboticS2R}, and self-driving vehicles \cite{RLS2R}.

\section{\uppercase{Method}}
\label{sec:Method}

\subsection{Training Environment}
\label{ssec:TrainingEnvironment}
Our simulation is based on {\em AirSim} \cite{airsim}, an open source simulation platform that is designed to experiment algorithms for various autonomous machines. To take advantage of the simulation, we invested our efforts in preparing a realistic simulation environment that will ease the sim-to-real process.
First, we designed a graphic model according to the prototype model that was made by the Technion Formula student team, prior the assembly of the original car. For this task, adjustments were made to the bone structure of the vehicle, along with the creation of the original materials and textures that were used to build the real car. This also included the drivetrain that served as a foundation for a realistic dynamic model. For this task, we cooperated with the engineers that were in charge of assembling the same car. Together, we adjusted mechanical components such as engine structure and gear, along with wheels and body properties, to have the simulated car behave similarly to the real one. To assure the correctness of the driving behavior, we measured lap times in simulation while using different track segments, and compared them to the real car performance. As for the environment itself, we modeled different types of traffic cones, according to the competition's regulations. To improve the variation of the data, we created a tool to efficiently draw varied track segments. Examples for the similarity between our simulated environment to the real one are shown in figure \ref{figure:collage}.

Another important stage for easing the process of moving from the simulation to the real world, was adjusting the simulated sensors to behave as close to the real ones. This was mainly composed of simulating the camera precisely, by placing it in the same position and adjusting technical properties accordingly, e.g., field of view and light sensitivity. For data recordings, we drove using a pro-level gaming steering wheel, so that for each captured image, we obtained an accurate steering angle, along with meta-data related to the current state of the driving, such as speed, throttle, brake and previous steering angle. 

\subsection{Model Architecture}
\label{ssec:ModelArchitecture}

\begin{figure}[t]
\centering
\includegraphics[width = 0.475\textwidth]{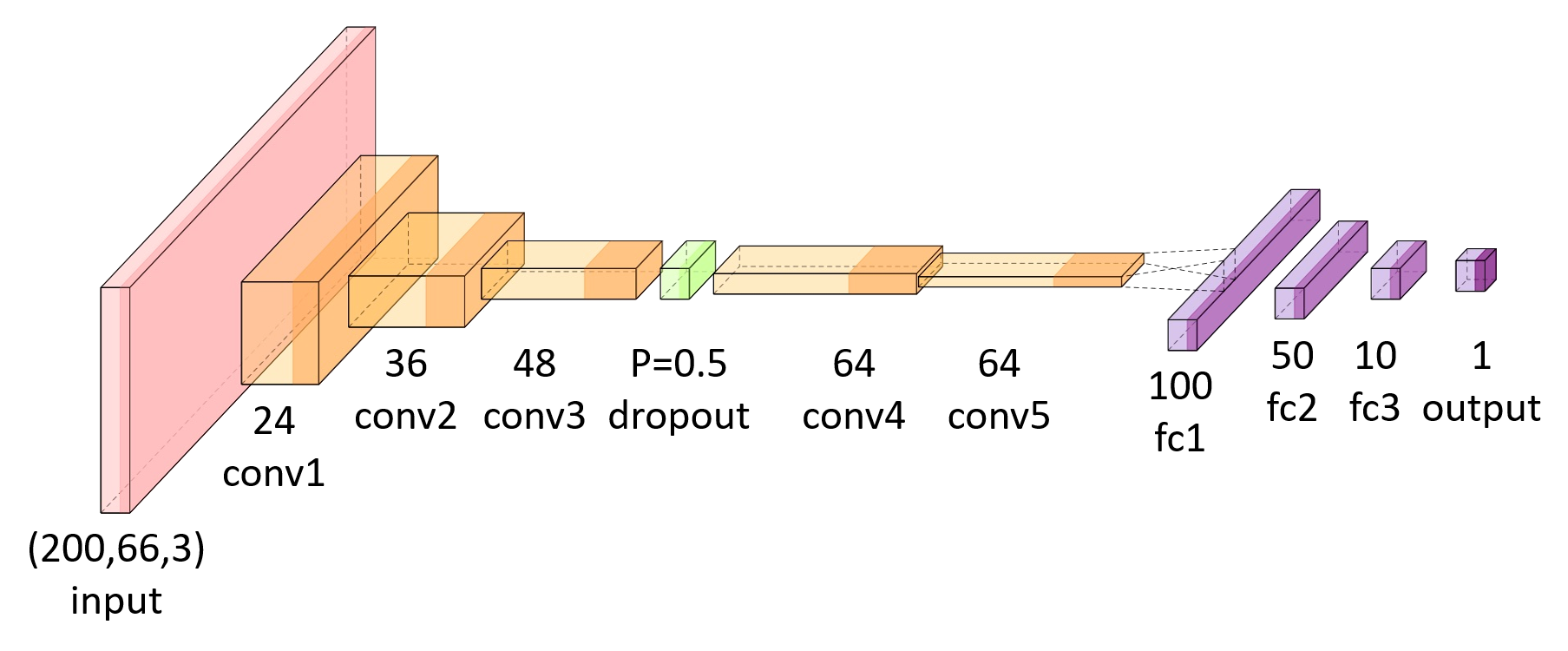}
\caption{Illustration of our modified PilotNet. We added a dropout layer (in green) in the middle of the convolutional block, ReLU as activation functions after each hidden layer and a Sigmoid activation function following the last layer.}
\label{figure:arch}
\end{figure}

Our DNN architecture is a modified version of PilotNet \cite{pilotnet}, an end-to-end network architecture that is designed to predict steering values for autonomous vehicles, based on a single captured image. Our modifications include using a Sigmoid layer as final activation, which allows us to predict $\hat{y} \in [0,1]$ as direct steering angle, instead of computing $\frac{1}{\hat{r}}$ where $\hat{r}$ is the circuit radius leading to the predicted steering. Also, we added ReLU activation following the hidden layers and a Dropout \cite{dropout} layer with a parameter $p=0.5$. A linear normalization of the image is expected to be performed by the users of the network prior to inference. A scheme of the final network is shown in figure \ref{figure:arch}.

\subsection{Augmentation and Data Enhancements}
\label{ssec:AugmentationandDataEnhancements}

Only a part of the recorded image is relevant in making the decision of the appropriate steering angle. Therefore, as shown in figure \ref{figure:diagram}, we crop a region of 200x66 pixels from the image that is fed into the network. To promote generalization, we used methods of data distortion and adjustments on sampled images. The difference in light exposure between real-world camera images to simulation recorded images invoked a need for manual change of brightness level. To address this, we duplicated the images in the pre-processing phase in variations of up to $40\%$ of the original brightness. We also used various settings of weather conditions when recording the data. These settings include cloud opacity and prevalence, light intensity and sun temperature. In addition, we used several post-process distortion methods on some of the sampled images. These included horizontal line contribution, horizontal distortion and lens flare reflection.

\begin{figure}[t]
\centering
\subfloat{\includegraphics[width = 0.095\textwidth]{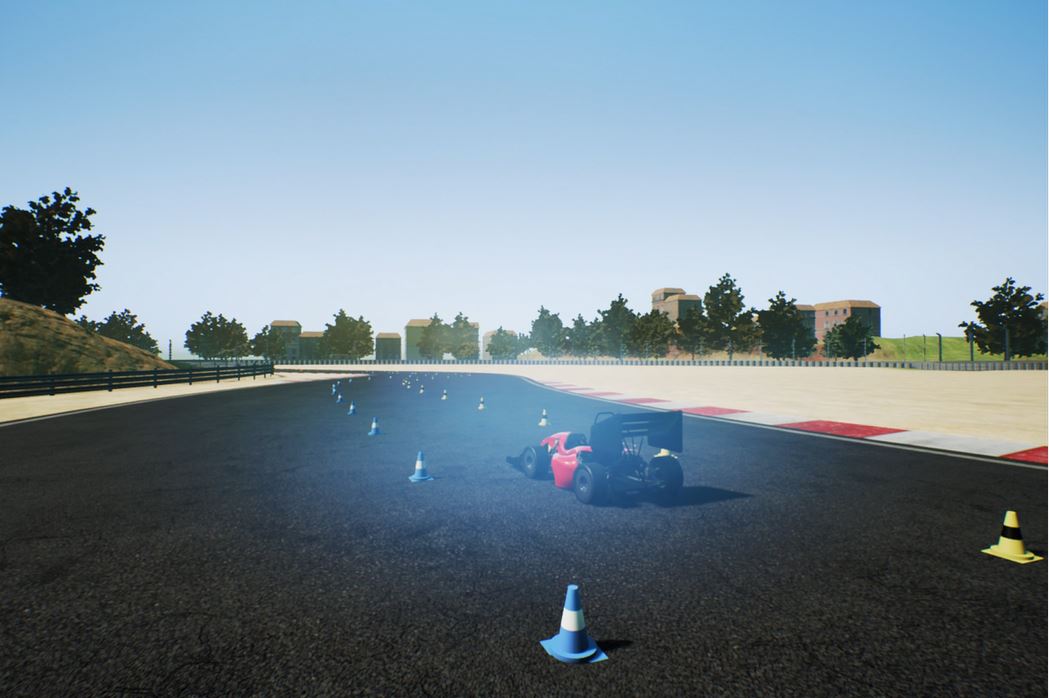}} 
\subfloat{\includegraphics[width = 0.095\textwidth]{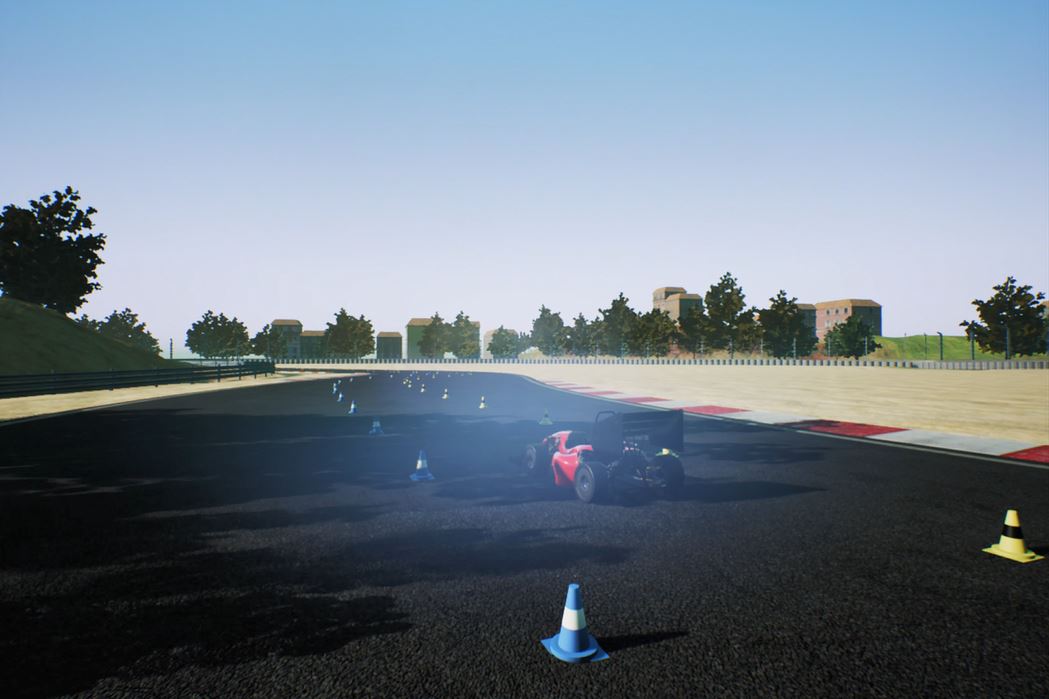}} 
\subfloat{\includegraphics[width = 0.095\textwidth]{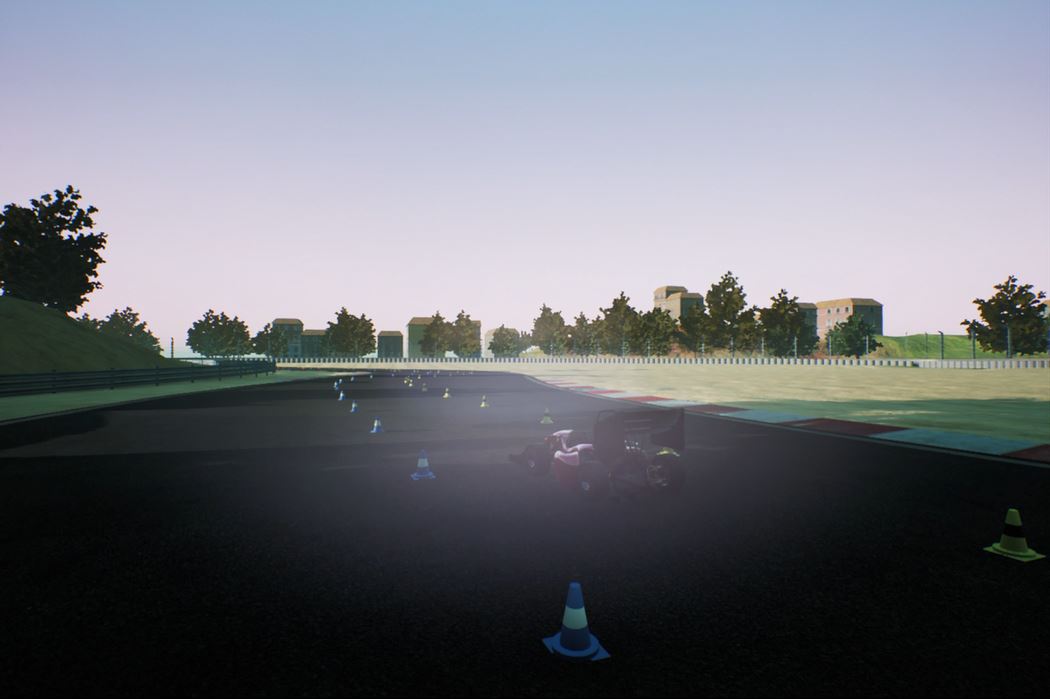}} 
\subfloat{\includegraphics[width = 0.095\textwidth]{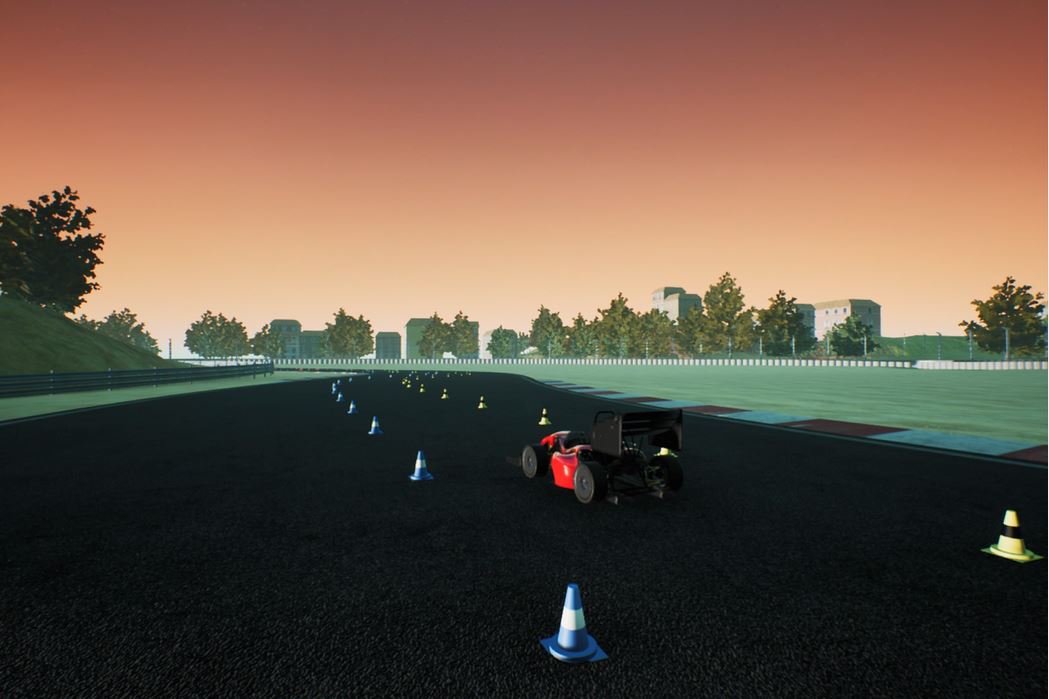}}
\subfloat{\includegraphics[width = 0.095\textwidth]{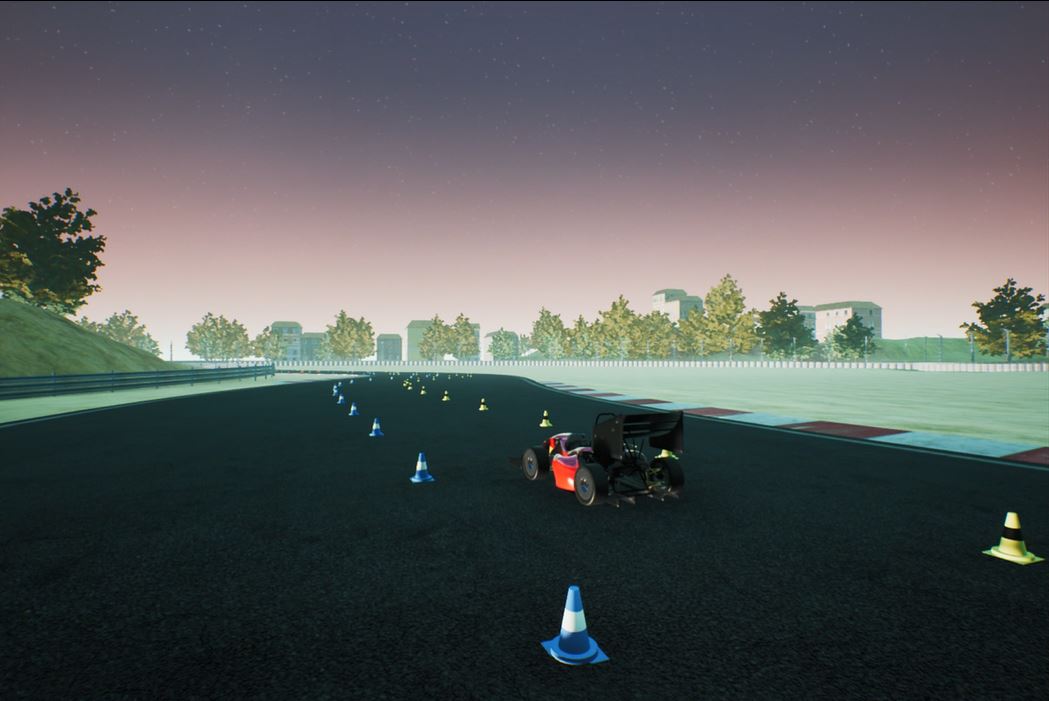}} \\ [-1.9ex]
\subfloat{\includegraphics[trim={0 5cm 0 0},clip,width = 0.234\textwidth]{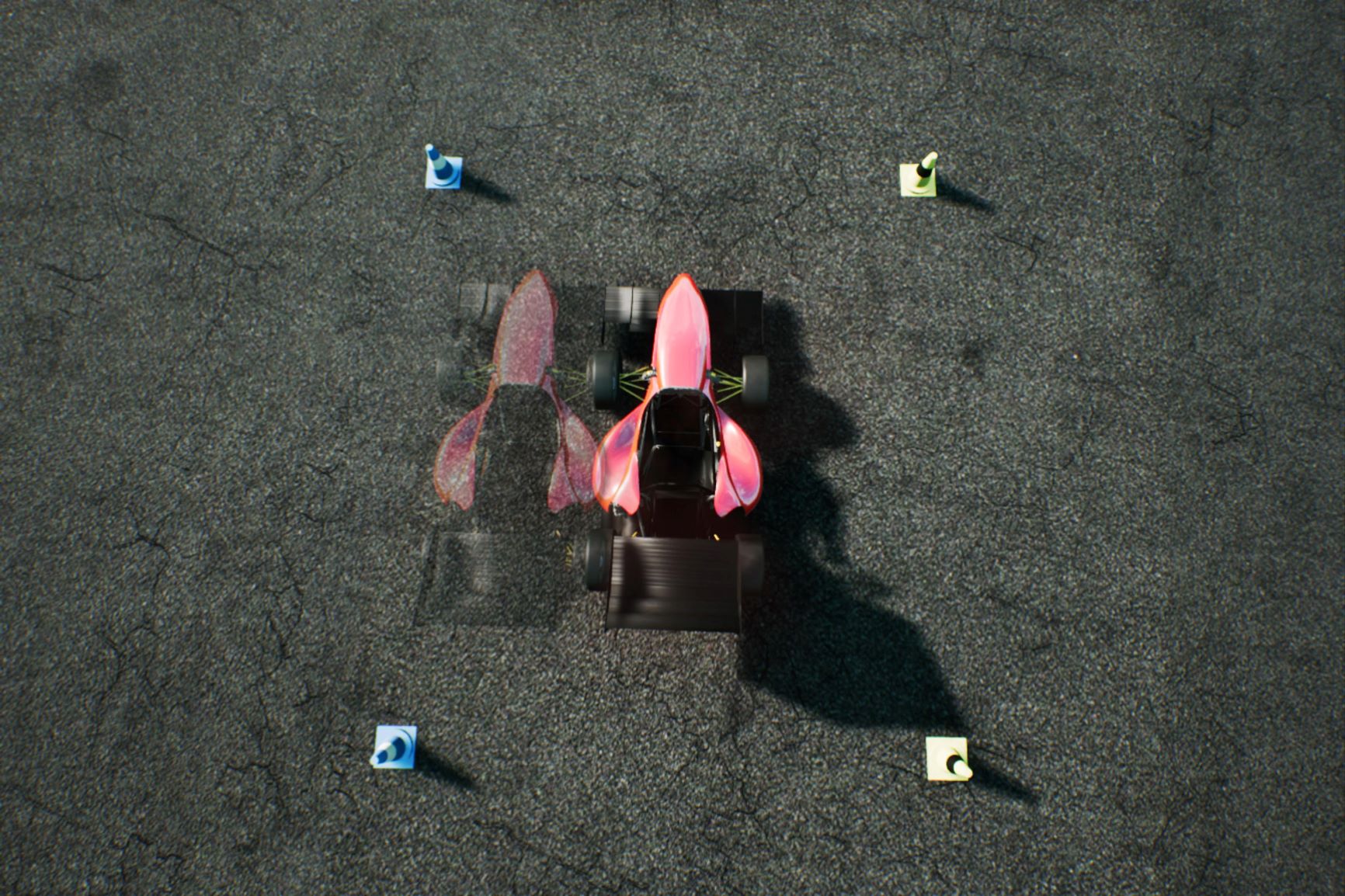}}\hspace{0.02cm}
\subfloat{\includegraphics[trim={0 5cm 0 0},clip,width = 0.234\textwidth]{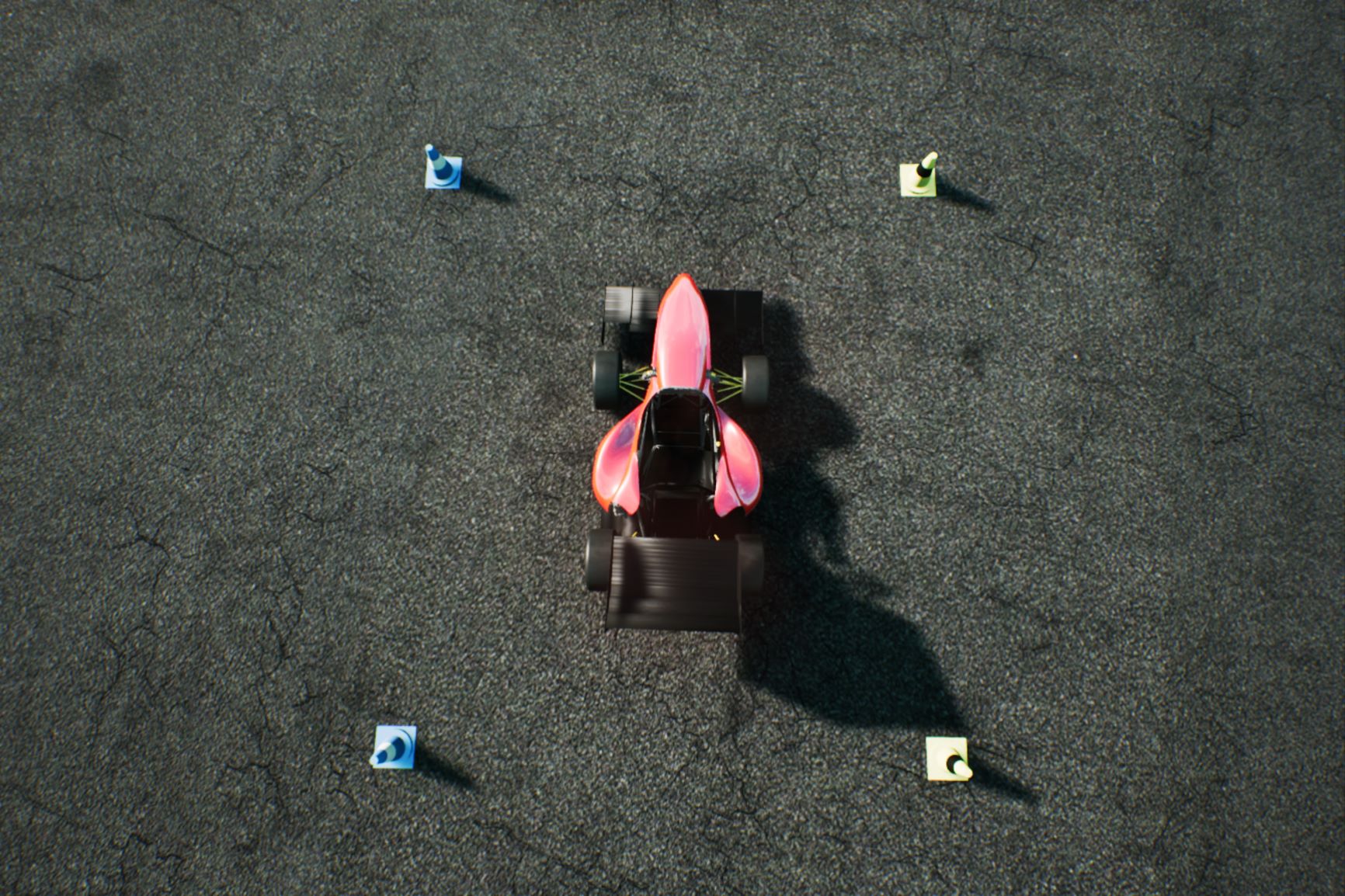}}
\caption{Illustration of our augmentation methods. Samples from CycleLight, a daylight cycle animation, demonstrating the effect of the time of the day on the environment (top). Shifted driving from a top view, illustrating the position of the car after shifting the camera one meter to the left (bottom).}
\label{figure:cyclelight-shifted}
\end{figure}

We propose a novel method we call {\it Shifted driving}. Based on the idea of clipping an image and post-processing the steering angle \cite{koppula2017learning}, we shift the camera's position along the width of the track and post-process the steering angle accordingly, as illustrated in figure \ref{figure:cyclelight-shifted}. Avoiding clipping an image and instead maintaining the entire region of interest, allows us to shift the camera's position more drastically. Our shifting distance in meters from the center, $d \in [-1.75,1.75]$, is normally distributed with zero mean.
Using this method improved our ability to face difficult situations that the car sometimes got into while self-driving and get back to the center of the track. It had made a great contribution to the ability of our model to maintain reliable driving, as will be discussed in sub-section \ref{ssec:SimulationExperiments}.
We also propose {\it CycleLight}, a novel method for introducing variation in recorded driving data in a simulated environment. {\it CycleLight}, as illustrated in figure \ref{figure:cyclelight-shifted}, is an animation of a day-light cycle in a changeable, potentially very short period of time, i.e., by turning this feature on during recording session, we could collect images from different hours of the day in just a few minutes, instead of manually recording a whole day in different conditions. The usage of this feature caused the model to become very robust to changes in shadowing conditions, daytime vs nighttime, ad signs, tire walls etc.

\begin{table*}[t]
\centering
\caption{Distribution for usage of the various techniques. Our augmentations contain methods to synthesize a given image, e.g., adding distortions and reflections, along with our unique methods to vary the recording position and animate a day-light cycle. Each percentage is related only to the mentioned technique, so overlaps between techniques are possible. The last technique shows that we used an almost equal amount of data from mostly straight tracks and from tracks containing many turns.}
\begin{tabular}{ccc}  
\toprule
Type  & Method & Percentage of usage \\
\midrule
\multirow{5}{*}{Augmentations}        & Horizontal line additions & 10.0\%          \\
                                      & Horizontal distortion     & 11.4\%          \\
                                      & Light reflection          & 9.9\%           \\
                                      & CycleLight                & 77.2\%          \\
                                      & Shifted driving           & 22.2\%          \\
\midrule
\multirow{2}{*}{Recording techniques} & Swerved driving           & 7.5\%           \\
                                      & Curvy/straight-line       & 46.1\% / 53.9\% \\
\bottomrule
\end{tabular}

\label{tab:methods}
\end{table*}

\subsection{Database and Recording Strategies}
\label{ssec:DatabaseandRecordingStrategies}

From different drivers, we chose three accurate and well-trained drivers to record driving sessions. The overall recording time for these drivers was approximately 220 minutes.
We built several virtual tracks. Almost half of the recorded data was taken from driving sessions on tracks which were mostly straight. For the other recordings, we used curvy tracks in which the number of left turns and the number of right turns were balanced. Since our competition regulations determine the distance between cones at each side of the track to be between 300 and 500 centimeters, we simulated different distances between cones during our training procedure.
Recordings were done by purposely using two driving styles, normative and {\it swerved}. Normative driving is the sensible way of keeping close to the center line of the road, while swerved driving means to constantly turn sharply from one edge of the road to the other. Without swerved driving style, our recordings were safe and steady, keeping close to the center of the road, not able to learn extreme situations. This method served the purpose of adding data samples of the car driving to the edge of the road and returning to the center.

\subsection{Real World Setup}
\label{ssec:RealWorldSetup}

The need to react quickly in racing competitions raises the importance of performing decision making in high frequencies. Such tasks, and especially tasks involving deep learning strategies, require an efficient computer that can manage parallel computations while operating under low-power constraints.
While our entire training procedure was based on Keras, an API that is not supported in some computer architectures, running our trained model on such computers required a transition of the model to TensorFlow framework. To increase the number of frames per second, we optimized our pre-processing phase. The optimization included cropping the relevant ROI from a given image prior to processing and replacing previous computations. The improvements in performance of such transition are described in section \ref{ssec:RealWorldExperiments} and are shown in table \ref{tab:optimization}. Our final program operates in 45 frames per second on average.

As for our camera, we used a single colored camera with a specific lens as the input device. The lens has a FOV (field of view) of 60\degree, like the simulated one, but it is vulnerable to changes in lighting exposure. Instead of manually adjusting the exposure, we used a software-based auto-exposure method. Then, we shipped our steering decisions using a serial port to the next stage, a micro-controller, that instructed the car's steering mechanism in combination with other dynamic constraints.

\section{\uppercase{Experiments}}
\label{sec:Experiments}

\noindent For testing our models, we used tracks with specifications that are dictated by the competition regulations. The simulation test-track featured driving at noon-time on mostly straight roads. The track included delicate turns, small hills, and shades. The car throttle value was determined by a function linear in the steering angle. One lap took 11 minutes of driving, with a maximum speed of approximately 25 km/h.
The metric used for benchmarking was the time passed from the moment the car started driving until it drove out of the track, as seen in table \ref{tab:arch} and discussed in section \ref{ssec:SimulationExperiments}.
The experiments composed of various changes in the initial PilotNet architecture, as demonstrated in figure \ref{figure:arch}, and shown in table \ref{tab:arch}.
In the real-world, the time of day for testing was mostly noon. We held two different experiments, one is a completely straight-line road and the second is a wave-like curved track.

During our benchmarking procedures, we tried different modifications to the network. 
Table \ref{tab:arch} shows the main points of change along with our training procedure from the perspective of data distribution and architecture. In contrary to what would seem to be the normative way of using a test set, we based our estimation of how good a model is by letting it drive through specific unseen tracks. This was done since there is more than one correct way to drive, thus having a test set would not necessarily generate proper feedback.
Addition of a Dropout layer produced an improvement in model performance, as mentioned in table \ref{tab:arch}. Also, a Sigmoid activation was added to the end of the network as an output normalizer.

After experiments with and without the car state, i.e., brakes, throttle, speed and previous steering as additional input, we observed over-fitting in the former case. e.g., after adding the swerved style data, the model was driving out of the center of the track more frequently and the performance worsened. The experiments described in table \ref{tab:arch} show improvement after removing the car state. We observed that it was better to discard these as inputs, despite their significance to the problem description.
After the addition of shifted driving, the final model performed exceptionally well on previously unseen tracks, as shown in table \ref{tab:arch}. Some of these tracks included objects beside the road. In all of these experiments, the car maintained an average speed of roughly 25 km per hour.

\subsection{Simulation Experiments}
\label{ssec:SimulationExperiments}
\begin{table}[t]
\centering
\caption{Comparison of our milestones throughout the experiments. We evaluate the model by measuring the time from the moment the car starts driving until it drives out of the track. When using a test set that is composed from a human recording, we observed that there is weak correlation between the evaluation loss and the driving behaviour. For example, adding the shifted driving method substantially enhanced the model's efficiency, an insight which we couldn't infer using a natural test set.}
\begin{tabular}{lc}  
\toprule
Modifications & Duration \\
\midrule
Original PilotNet           & 20 seconds \\
Dropout, Leaky ReLU         & 3 minutes  \\
Addition of swerved data    & 5 minutes  \\
Removal of car state        & 6 minutes  \\
Shifted driving             & 3 hours    \\
\bottomrule
\end{tabular}
\label{tab:arch}
\end{table}

\begin{figure}[t]
\centering
\includegraphics[width = 0.42\textwidth]{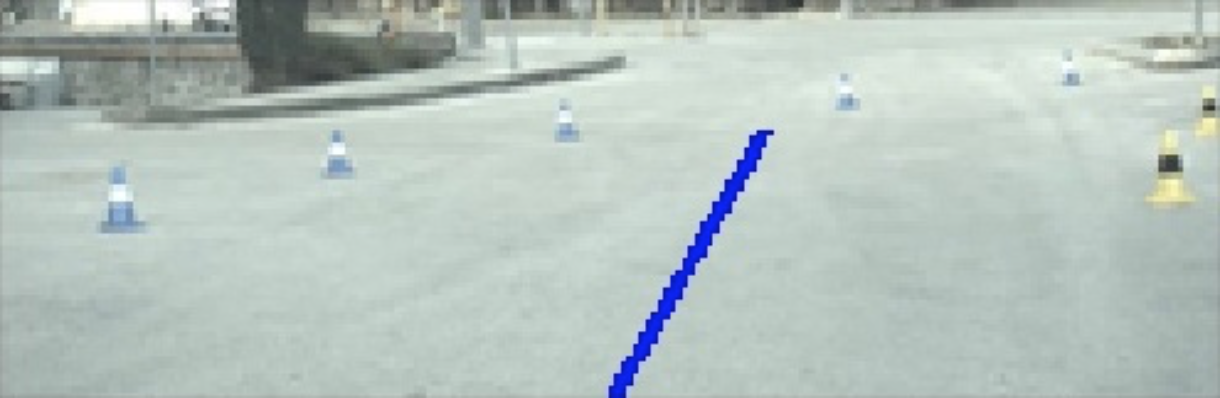}
\caption{A sample taken from a recorded video in a parking lot. The predicted steering angle is marked with a blue line on the image. At this stage, the trained model from the simulation is tested on the video, without any conversions.}
\label{figure:labeledpic}
\end{figure}

\subsection{Transition from Simulation to Real World}
\label{ssec:TransitionfromSimulationtoRealWorld}

To ensure the precision of the model in the real world, we tested it offline on given videos of driving on a closed-off road track. Due to the lack of labels in our videos, we conducted these experiments using the following procedure: we marked the predicted steering angles on the videos using a drawn line, as shown in figure \ref{figure:labeledpic}. We were assisted by experts from the dynamic team of the car who tested the videos and helped us to realize if the predicted steering angles had the expected directions.

\begin{table}[t]
\centering
\caption{Execution of one hundred samples in both computers, running on three different code implementations. On the left column is a desktop computer containing Nvidia Geforce GTX1080, while on the right is a Nvidia Jetson TX2. The results are presented in average execution time per iteration, in seconds. Due to lack of support, we couldn't evaluate Keras inference on TX2.}
\begin{tabular}{l|cc}  
\toprule
 & x86 & TX2 \\
\midrule
Keras inference         & 0.0026 sec & -  \\
TensorFlow inference    & 0.0351 sec & 0.4174 sec \\
Optimized inference  & 0.0025 sec & 0.0176 sec \\
\bottomrule
\end{tabular}
\label{tab:optimization}
\end{table}

As mentioned before, we needed to convert our model to TensorFlow framework and optimize response time. To check the validity of the transition, we tested both Keras and TensorFlow models on given videos. We compared the inference results of both models, assuming they should produce the same outputs. Also, we checked the inference time in both cases. We took one hundred samples and let both computers execute inference on them. As shown in table \ref{tab:optimization}, we assured that our optimized code can execute sufficient inference frequencies on our designated computer.

\subsection{Real World Experiments}
\label{ssec:RealWorldExperiments}

For the sake of simplicity, we began by using a “clean” environment, i.e., one that included only our Jetson TX2 computer, a camera, a micro-controller and traffic cones in a lab, despite many differences from an outdoor road scene. This helped us in avoiding the interruptions of mechanical issues within the car. We set the experiments by placing traffic cones in the lab’s open space in different positions, to create various indoor track segments.

\begin{figure}[t]
\centering
\includegraphics[width = 0.4\textwidth]{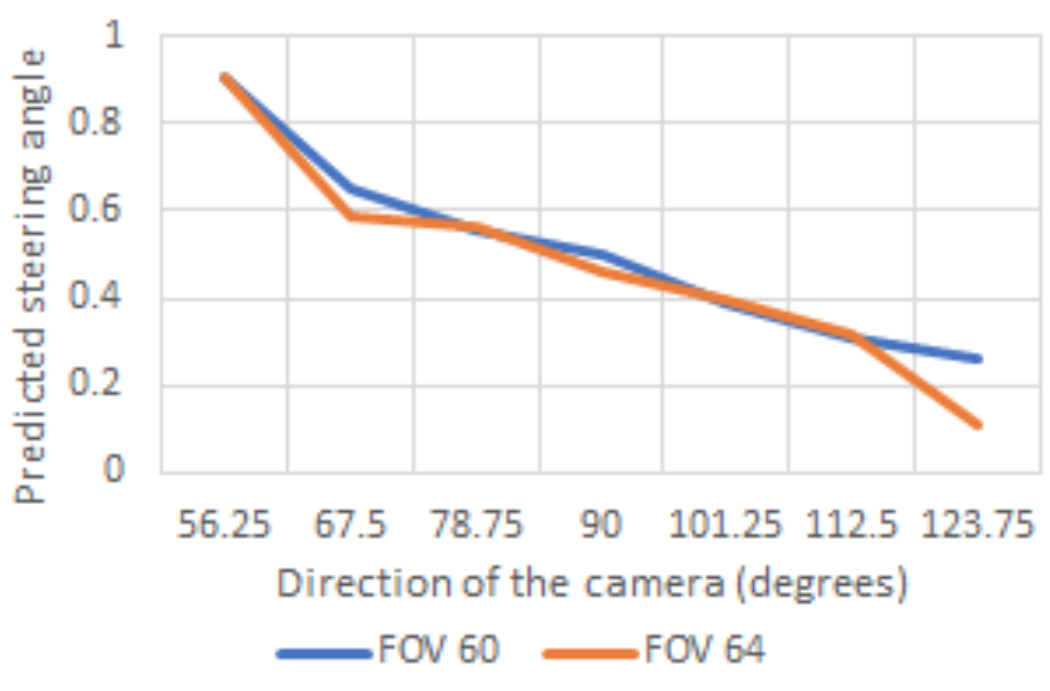}
\caption{Predicted steering angle as a function of placement direction of the cameras. The graph shows the difference between two chosen cameras.}
\label{figure:fovgraph}
\end{figure}

We investigated the effects of changing the camera's parameters and realized that FOV, aperture diameter, shutter speed and color scheme influenced the most. To adjust the attributes of the camera to align with the simulated one, we calculated $f$, the focal length of the lens which depends on $x$, the FOV of the simulated camera, and on $d$, the diameter of the sensor:
\begin{equation}
f = \frac{d}{2\tan(\frac{x}{2})}
\end{equation}
Where in our case $d$ equals to 11.345 mm and $x$ equals to 60\degree.
To prove the necessity in using the same parameters of the camera in real-world like the simulated one, we conducted an experiment in a lab using two different cameras. One that had a FOV of 60\degree, like the simulated one and the other with a FOV of 64\degree. We placed traffic cones in two rows in front of the cameras and installed both cameras in the same position, in several rotations. We were expecting to find a meaningful pattern with respect to the direction values. The results showed us a less spiked behavior in the slope of the graph for the camera with the same FOV as the simulated one, which means a smooth tendency in the difference between predicted steering angles. This is shown in figure \ref{figure:fovgraph}.

To distinguish between the execution of the program to the mechanical work of the controller, we tested the connection to the controller without using the actual car. This experiment was performed using an Arduino module as a micro-controller receiving the output from our Jetson TX2, and presenting the predicted value using a LED light bulb.
Our next step was the outdoor scene. At first, the experiments were held statically in a closed-off road scene. We placed a work-station in different positions, consisting of the Jetson TX2 and the camera, which were placed on a track marked by traffic cones. The camera’s height was adjusted to its original height on the actual car. This was the first time we encountered sensitive exposure, this led us to correct the exposure using programmatic auto-exposure solutions.

The last experiments took place on a parking lot and finally on an obstacle-free road segment, using the actual car. The speed was increased progressively. To cope with the operations of mechanical systems, such as clutch and throttle control, we were forced to test the program’s performance with slow speeds. The best way to do so was without ignition while turning on the steering system alone.
One of the main problems was a delay in steering execution. The dynamic model of the steering mechanism was under construction until our experiments phase, hence the simulated model did not consider the mechanical delay times accurately. Sending a steering value from the model to the controller took negligible time while mechanically changing the direction of the wheels took two times longer than expected. Thus, executing the steering angle instruction was slow, the car consistently drove out of track and the car’s attempts to get back on track were unsuccessful. After a trial and error procedure, we managed to identify the optimal driving speed, which was 15-20 km/h, limited by mechanical constrains. 

Also, we encountered dazzling when driving in a specific time of the day, in different directions. Such phenomena resulted in unwanted anomalies during inference, which often led us to abort the driving unwillingly. One way of dealing with such a problem is to use multiple cameras or additional camera filters. Also, the usage of distance estimators, e.g., Lidar or depth cameras, can help overcome such distortions in real world images.
The driving results in real-world were highly successful with the model trained on data gathered from simulation only. The car drove better than expected and was able to finish track segments of 100-200 meters long. 

\begin{figure}[t]
\centering
\subfloat{\includegraphics[width = 0.22\textwidth]{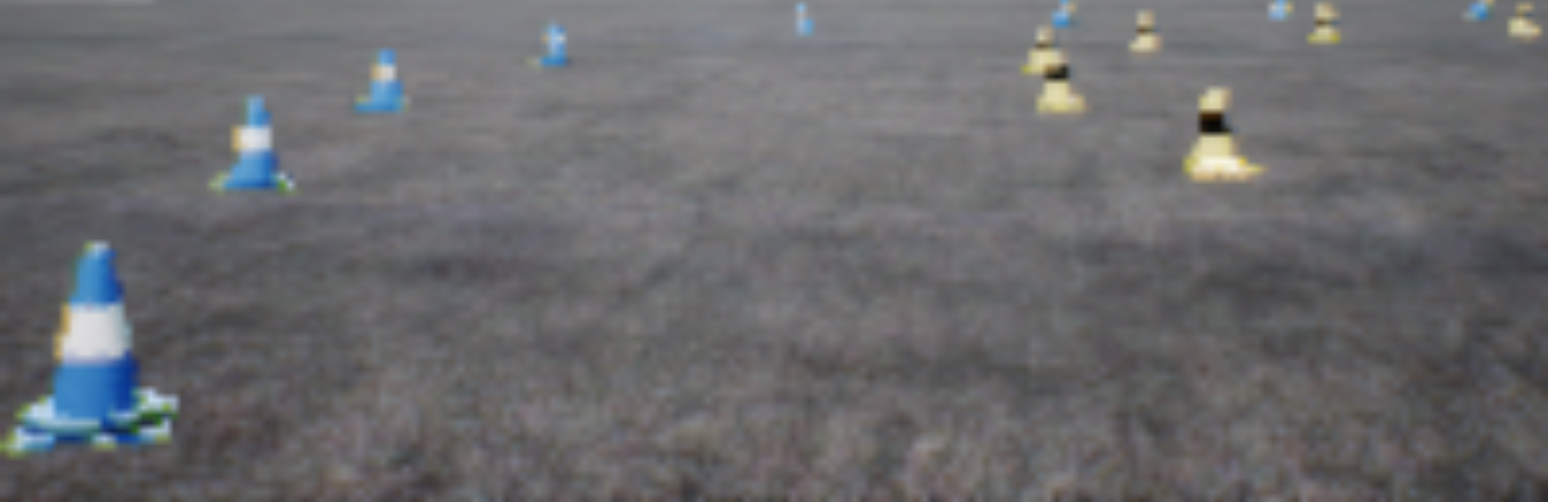}}\hspace{0.02cm}
\subfloat{\includegraphics[width = 0.22\textwidth]{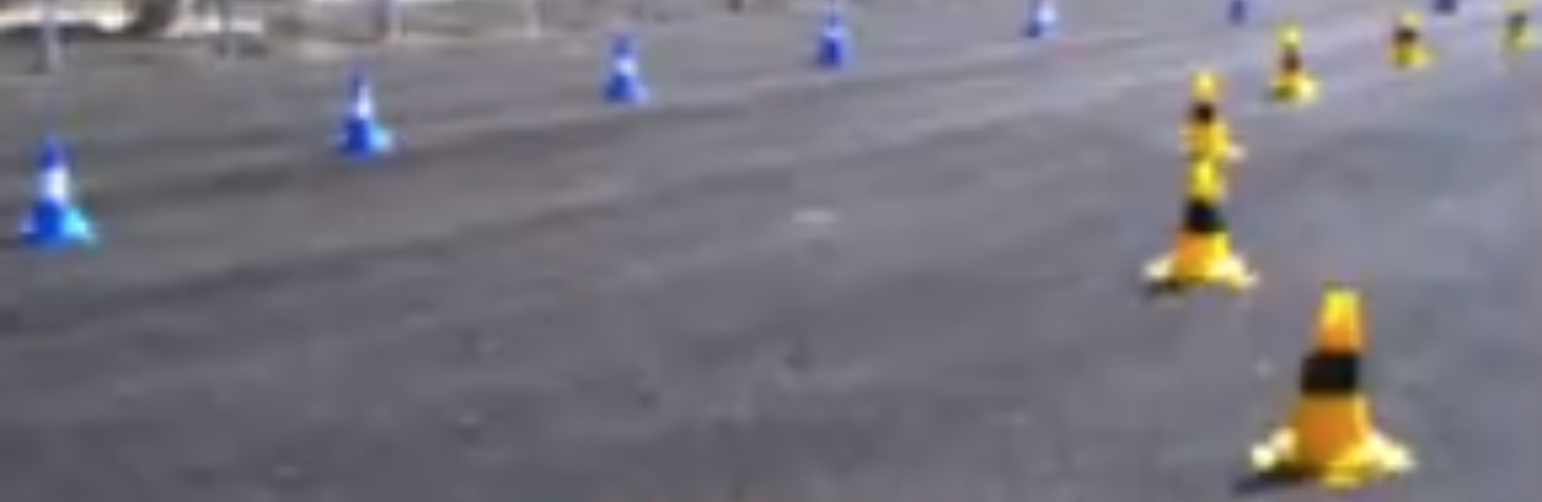}}
\caption{Comparison between sampled images taken from our simulated camera (left) and our real world camera (right).}
\label{figure:track}
\end{figure}

\section{\uppercase{Discussion}}
\label{sec:Discussion}

\noindent A prominent aspect of what makes our training procedure interesting is the fact that no real-world data was used, this greatly simplified the data gathering process. Augmentation and recording techniques that were used to compensate for the lack of real-world data, proved to be crucial for the successful transition of the model to the actual car. Also noteworthy is the fact that no depth related sensor was used in any part of our work, moreover, we managed to implement a well-performing algorithm using only a single camera as our input, an uncommon practice in the self-driving field. Finally, we have also shown that such algorithms can run under low-power constraints in a sufficiently high enough throughput.

At a broader sense than previously discussed, the concepts we used during our training procedure could be deployed when attempting a variety of different tasks. One could think of utilizing CycleLight to varying lighting conditions when training off-road vehicles to maneuver between obstacles, or rethinking the idea of shifted driving to spread out on a two-dimensional space when training trajectory planning for air crafts. On such cases, we believe that simulation-based training would greatly simplify the process.

\section*{\uppercase{Acknowledgements}}

\noindent The research was sopported by the Intelligent Systems Lab (ISL) and the Center for Graphics and Geometric Computing (CGGC), CS faculty, Technion. The Formula SAE car was supplied by the Technion Formula Student Team.

\bibliographystyle{apalike}
{\small
\bibliography{main}}

\begin{thebibliography}{}

\bibitem[Bast et~al., 2016]{BAS15}
Bast, H., Delling, D., Goldberg, A.~V., M{\"u}ller-Hannemann, M., Pajor, T.,
  Sanders, P., Wagner, D., and Werneck, R.~F. (2016).
\newblock Route planning in transportation networks.

\bibitem[Bojarski et~al., 2016]{pilotnet}
Bojarski, M., Del~Testa, D., Dworakowski, D., Firner, B., Flepp, B., Goyal, P.,
  Jackel, L.~D., Monfort, M., Muller, U., Zhang, J., et~al. (2016).
\newblock End to end learning for self-driving cars.
\newblock {\em arXiv preprint arXiv:1604.07316}.

\bibitem[Brubaker et~al., 2016]{BRU16}
Brubaker, M.~A., Geiger, A., and Urtasun, R. (2016).
\newblock Map-based probabilistic visual self-localization.
\newblock {\em IEEE Trans. Pattern Anal. Mach. Intell.}, 38(4):652--665.

\bibitem[Chebotar et~al., 2018]{S2R2}
Chebotar, Y., Handa, A., Makoviychuk, V., Macklin, M., Issac, J., Ratliff,
  N.~D., and Fox, D. (2018).
\newblock Closing the sim-to-real loop: Adapting simulation randomization with
  real world experience.
\newblock {\em CoRR}, abs/1810.05687.

\bibitem[Chen et~al., 2017]{DEEPLAB}
Chen, L., Papandreou, G., Kokkinos, I., Murphy, K., and Yuille, A.~L. (2017).
\newblock Deeplab: Semantic image segmentation with deep convolutional nets,
  atrous convolution, and fully connected crfs.
\newblock {\em CoRR, 2017 IEEE Transactions on Pattern Analysis and Machine
  Intelligence}, abs/1606.00915.

\bibitem[Codevilla et~al., 2018]{codevilla2018end}
Codevilla, F., Miiller, M., L{\'o}pez, A., Koltun, V., and Dosovitskiy, A.
  (2018).
\newblock End-to-end driving via conditional imitation learning.
\newblock In {\em 2018 IEEE International Conference on Robotics and Automation
  (ICRA)}, pages 1--9. IEEE.

\bibitem[Glassner et~al., 2019]{E2E1}
Glassner, Y., Gispan, L., Ayash, A., and Shohet, T.~F. (2019).
\newblock Closing the gap towards end-to-end autonomous vehicle system.
\newblock {\em CoRR}, abs/1901.00114.

\bibitem[James et~al., 2018]{S2R1}
James, S., Wohlhart, P., Kalakrishnan, M., Kalashnikov, D., Irpan, A., Ibarz,
  J., Levine, S., Hadsell, R., and Bousmalis, K. (2018).
\newblock Sim-to-real via sim-to-sim: Data-efficient robotic grasping via
  randomized-to-canonical adaptation networks.
\newblock {\em CoRR}, abs/1812.07252.

\bibitem[Kiran et~al., 2018]{BRavi}
Kiran, B.~R., Rold{\~{a}}o, L., Irastorza, B., Verastegui, R., S{\"{u}}ss, S.,
  Yogamani, S., Talpaert, V., Lepoutre, A., and Trehard, G. (2018).
\newblock Real-time dynamic object detection for autonomous driving using prior
  3d-maps.
\newblock {\em CoRR, ECCV 2018}.

\bibitem[Koppula, 2017]{koppula2017learning}
Koppula, S. (2017).
\newblock Learning a cnn-based end-to-end controller for a formula sae racecar.
\newblock {\em arXiv preprint arXiv:1708.02215}.

\bibitem[Levinson and Thrun, 2010]{RobustVL}
Levinson, J. and Thrun, S. (2010).
\newblock Robust vehicle localization in urban environments using probabilistic
  maps.
\newblock {\em 2010 IEEE International Conference on Robotics and Automation},
  pages 4372--4378.

\bibitem[Li et~al., 2019]{Peili}
Li, P., Chen, X., and Shen, S. (2019).
\newblock Stereo {R-CNN} based 3d object detection for autonomous driving.
\newblock {\em CoRR}, abs/1902.09738.

\bibitem[Mehta et~al., 2018]{E2E2}
Mehta, A., Subramanian, A., and Subramanian, A. (2018).
\newblock Learning end-to-end autonomous driving using guided auxiliary
  supervision.
\newblock {\em CoRR}, abs/1808.10393.

\bibitem[Peng et~al., 2017]{RoboticS2R}
Peng, X.~B., Andrychowicz, M., Zaremba, W., and Abbeel, P. (2017).
\newblock Sim-to-real transfer of robotic control with dynamics randomization.
\newblock {\em CoRR}, abs/1710.06537.

\bibitem[Redmon et~al., 2015]{YOLO}
Redmon, J., Divvala, S.~K., Girshick, R.~B., and Farhadi, A. (2015).
\newblock You only look once: Unified, real-time object detection.
\newblock {\em CoRR}, abs/1506.02640.

\bibitem[Shah et~al., 2018]{airsim}
Shah, S., Dey, D., Lovett, C., and Kapoor, A. (2018).
\newblock Airsim: High-fidelity visual and physical simulation for autonomous
  vehicles.
\newblock In {\em Field and service robotics}, pages 621--635. Springer.

\bibitem[Srivastava et~al., 2014]{dropout}
Srivastava, N., Hinton, G., Krizhevsky, A., Sutskever, I., and Salakhutdinov,
  R. (2014).
\newblock Dropout: a simple way to prevent neural networks from overfitting.
\newblock {\em The Journal of Machine Learning Research}, 15(1):1929--1958.

\bibitem[Tan et~al., 2018]{QuadS2R}
Tan, J., Zhang, T., Coumans, E., Iscen, A., Bai, Y., Hafner, D., Bohez, S., and
  Vanhoucke, V. (2018).
\newblock Sim-to-real: Learning agile locomotion for quadruped robots.
\newblock {\em CoRR}, abs/1804.10332.

\bibitem[Urmson et~al., 2008]{darpa}
Urmson, C., Anhalt, J., Bagnell, D., Baker, C., Bittner, R., Clark, M., Dolan,
  J., Duggins, D., Galatali, T., Geyer, C., et~al. (2008).
\newblock Autonomous driving in urban environments: Boss and the urban
  challenge.
\newblock {\em Journal of Field Robotics}, 25(8):425--466.

\bibitem[Valls et~al., 2018]{slam}
Valls, M.~I., Hendrikx, H.~F., Reijgwart, V.~J., Meier, F.~V., Sa, I.,
  Dub{\'e}, R., Gawel, A., B{\"u}rki, M., and Siegwart, R. (2018).
\newblock Design of an autonomous racecar: Perception, state estimation and
  system integration.
\newblock In {\em 2018 IEEE International Conference on Robotics and Automation
  (ICRA)}, pages 2048--2055. IEEE.

\bibitem[Xu et~al., 2017]{XU17}
Xu, Y., John, V., Mita, S., Tehrani, H., Ishimaru, K., and Nishino, S. (2017).
\newblock 3d point cloud map based vehicle localization using stereo camera.
\newblock {\em 2017 IEEE Intelligent Vehicles Symposium (IV)}.

\bibitem[You et~al., 2017]{RLS2R}
You, Y., Pan, X., Wang, Z., and Lu, C. (2017).
\newblock Virtual to real reinforcement learning for autonomous driving.
\newblock {\em BNVC}.

\end{thebibliography}

\end{document}